%% file: nips_2018.tex
\documentclass{article}

\usepackage[final]{nips_2018}

\usepackage[]{algorithm, algorithmic}

\usepackage[utf8]{inputenc} 
\usepackage[T1]{fontenc}    
\usepackage{hyperref}       
\usepackage{url}            
\usepackage{booktabs}       
\usepackage{amsfonts}       
\usepackage{nicefrac}       
\usepackage{microtype}      

\usepackage{microtype}
\usepackage{graphicx}
\usepackage{subfigure}
\usepackage{booktabs} 
\usepackage[utf8]{inputenc} 
\usepackage[T1]{fontenc}    
\usepackage{url}            
\usepackage{booktabs}       
\usepackage{amsfonts}       
\usepackage{nicefrac}       
\usepackage{microtype}      
\usepackage{amsmath, amsthm}

\usepackage{color}

\newcommand{\E}{\mathbb{E}}

\title{Robust Classification of Financial Risk}

\author{
  Suproteem K.~Sarkar\thanks{\noindent Equal Contribution} \\
  Department of Computer Science\\
  Harvard University\\
  Cambridge, MA 02138 \\
  \texttt{suproteemsarkar@g.harvard.edu} \\
    \And
     Kojin Oshiba\textsuperscript{*} \\
    Department of Computer Science\\
  Harvard University\\
  Cambridge, MA 02138 \\
    \texttt{kojinoshiba@college.harvard.edu} \\
   \AND
  Daniel Giebisch\textsuperscript{*} \\
   Department of Computer Science\\
  Harvard University\\
  Cambridge, MA 02138 \\
   \texttt{danielgiebisch@college.harvard.edu} \\
   \And
   Yaron Singer \\
   Department of Computer Science\\
  Harvard University\\
  Cambridge, MA 02138 \\
   \texttt{yaron@seas.harvard.edu} \\
}

\begin{document}

\maketitle

\begin{abstract}
Algorithms are increasingly common components of high-impact decision-making, and a
growing body of literature on adversarial examples in laboratory settings indicates that standard machine learning models are not robust.
This suggests that real-world systems are also susceptible to manipulation or misclassification, which especially poses
a challenge to machine learning models used in financial services. 
We use the loan grade classification problem to explore how machine learning models are sensitive to small changes in user-reported data, using adversarial attacks documented in the literature and an original, domain-specific attack. 
Our work shows that a robust optimization algorithm can build models for financial services that are resistant to misclassification on perturbations. To the best of our knowledge, this is the first study of adversarial attacks and defenses for deep learning in financial services.
\end{abstract}

\input{intro}

\input{methods}

\input{attacks}
\input{classifier}
\medskip

\small

\bibliography{nips}
\bibliographystyle{unsrt}

\end{document}

%% file: intro.tex
\section{Introduction}

In this paper we explore the robustness of algorithmic decision-making systems that rely on machine learning.  In recent years machine learning has become an integral component of financial decision-making systems \cite{bao2017deep,alberg2017improving,sirignano2016deep,smalter2017macroeconomic}, and is often used to predict loan defaults and determine interest rates.  Although deep learning classifiers are capable of producing remarkably accurate predictions, they are susceptible to manipulation. For image classification, recent research has shown that state-of-the-art deep learning models can be fooled to output false image labels by adding small---albeit carefully constructed---noise to the image~\cite{szegedy2013intriguing, madry2017towards, papernot2016limitations}.  Consequently, a great deal of effort in the machine learning community is placed on developing image classifiers that are robust to adversarial manipulations.

For financial applications, like the prediction of loan default to determine interest rates, individuals may be incentivized to manipulate data if small perturbations in the input data can lead to improved interest rates.  Financial decision-making services must therefore be robust to such perturbations. 
The potential for misclassification in these models poses two key questions:
\begin{enumerate}
    \item \emph{Do small perturbations to input data lead to misclassification?}
    \item \emph{Can robust classification help machine learning models used in financial services perform better on perturbed examples?}
\end{enumerate}

\begin{figure}[t]
\centering
\includegraphics[width=0.8\linewidth]{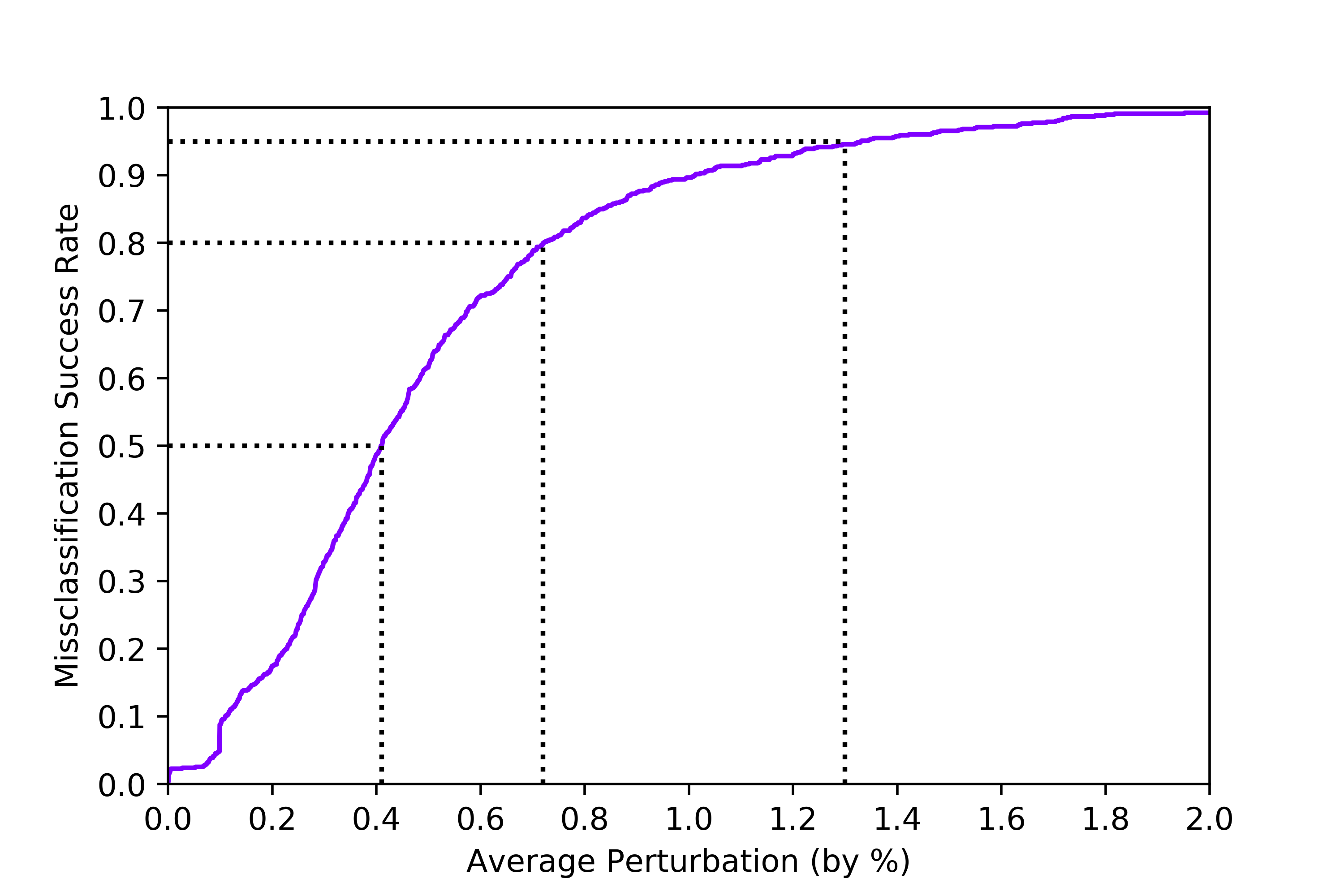}
\caption{Misclassification of perturbed loan profiles as a function of average perturbation (by percent). Note that a 1.3\% average change per feature leads to a misclassification on 95\% of examples.}
\label{fig:intro_fig}
\end{figure}

In this paper we address these questions and provide affirmative responses to both.  We show minor misreporting of certain features can cause individuals to improve their interest rates.  However, we show that robust optimization techniques can be used to train a classifier that is robust to such manipulation.  More concretely, we use a dataset of loan applicant information and loan grades from LendingClub \cite{lendingstats} as a case study for demonstrating how real-world algorithmic systems are used in financial services, and obtain the following results:
\begin{itemize}
\item We first show that that small perturbations on our high-accuracy loan grade prediction model are successful, with average perturbations of $1.3\%$ per feature leading to misclassification on 95\% of inputs (Figure~\ref{fig:intro_fig}); 
\item We build a robust classifier on models that is adversarially trained across perturbation types. We use common adversarial attacks from the literature, and propose our own attack to model a low-resource information manipulator. We show that our robust classifier achieves an increase in accuracy of 20\% against the benchmark when the model is evaluated on the minmax objective. 
\end{itemize}

\subsection{Paper organization}

In Section 2, we discuss our methods of loan grade classification and adversarial perturbation, and describe our robust optimization algorithm. In Section 3, we evaluate misclassification efficacy across perturbation types. Finally, in Section 4, we present the results of training with our robust classification algorithm.

%% file: methods.tex
\section{Methods}
 In this section, we introduce the dataset, perturbations, and the robust optimziation algorithm used throughout the paper.

\subsection{LendingClub Data} \label{loan}
We use loan application data from a set of 421,095 applications submitted to LendingClub in 2015 \cite{lendingstats}. LendingClub is an online peer-to-peer lending platform that algorithmically assigns loan categories based on application data. LendingClub and other online loan services base their loan categories on several user-inputted features in order to achieve more accurate credit assessment than traditional loan services---including banks, which mainly use FICO credit scores. Each applicant is assigned a grade from A to G by the platform, where A assigns the lowest interest rates (between 5-8\%) and G assigns the highest rates (between (25-29\%). A change to an adjacent loan grade can entail a 2-5\% change in interest rate. LendingClub also assigns applicants to one of five subgrades per category (for example, A is divided between A1-A5).
Our analysis focuses on differences between the seven large loan grades to show the significant magnitude by which adversarial perturbation can affect expected results.

We train a neural network as a proxy for the classification system built by LendingClub, as we do not have access to the service's own proprietary model.  Specifically, we train a neural network to predict loan grade from 44 continuous features including FICO score, debt-to-income ratio, and loan-to-income ratio, and 6 discrete features, including state and filing status. Our network uses 2 dense ReLU layers of size 100 and 60, with a dropout of 0.2 and a final softmax layer to define predictions for 7 possible loan grades. This architecture is based off Andersen's architecture for approaching this classification problem \cite{andersen_2018}. We achieve a test accuracy of $94.46\%$.

\subsection{Pertubation Types} \label{perturbations}
We evaluate a number of adversarial methods on the heterogeneous feature space of our model, targetting the 44 continuous features. 
Adversarial attacks perturb an input up to a designated noise budget, under the intuition that with a sufficiently small budget any perturbation in an input is "meaningless" in terms of actually changing the ground truth classification of the input. Furthermore, we standardize each feature in the training and perturbation procedures before applying the perturbation. This is because unlike laboratory datasets used in the robustness literature (including images), the features relevant to loan grade prediction are not drawn from the same supports. 

\textbf{Fast Gradient Sign Method (FGSM)} perturbs an input in the direction of the gradient of the loss function of the model with respect to the input \cite{goodfellow2014explaining}. The size of the perturbation of each feature is a constant parameter $\epsilon$.
Given a loss function $L(\theta, x, y)$ with parameters $\theta$, input $x$, and label $y$, FGSM produces the perturbation 

\vspace*{-1.0em}
\begin{equation*}
x^* = x + \epsilon\operatorname{sign}(\nabla_{x} L(\theta, x, y)).
\end{equation*}
\vspace*{-1.0em}

\textbf{Projected Gradient Descent (PGD)} is a similar process to FGSM, but descends the actual gradient, as opposed to the sign \cite{madry2017towards}. Additionally, the gradient is recomputed at every step. The maximum perturbation for each iteration is clipped by a parameter $\epsilon$. PGD iteratively produces perturbed examples $x^*_i$ (up to the maximum step $x^*_N$), following the procedure

\vspace*{-1.5em}
\begin{eqnarray*}
x^*_0 &=& x, \\
x^*_i &=& \operatorname{clip}_{\epsilon}(x^*_{i - 1} + \epsilon
(\nabla_{x^*_{i - 1}} L(\theta, x^*_{i - 1}, y))).
\end{eqnarray*}
\vspace*{-1.5em}

\textbf{Jacobian-based Saliency Mapping Attack (JSMA)} isolates and perturbs the features with highest impact on the loss of a target label \cite{papernot2016limitations}. JSMA greedily increases the probability of the target label while decreasing the probabilities of all other labels. With $F_j(x)$ as the output of a neural network for class $j$, a saliency map is defined as

\vspace*{-1.3em}
\begin{align*}
S(x,t)[i] = 
\begin{cases} 
0, \text{ if }\ \frac{\partial F_t(x)}{\partial x_i} < 0
\text{ or } \sum_{j \neq t} \frac{\partial F_j(x)}{\partial x_i} > 0 \\  
(\frac{\partial F_t(x)}{\partial x_i})|\sum_{j \neq t} \frac{\partial F_j(x)}{\partial x_i}|,
\text{ otherwise} 
\end{cases}
\end{align*}

JSMA perturbs the features with the highest impact on salience by a constant perturbation $\epsilon$.

\textbf{Max Salience Attack (MSA)} is an attack we propose. This attack mimics self-report manipulation, where a loan applicant might change a small number of inputs to try to affect loan grade. Specifically, we perturbed each feature individually by a set percent increase or decrease, and found which of these features produced the highest rate of successful misclassifications. We then perturbed one and two of these most salient features (1-MSA, 2-MSA) by a fixed percent change. While these attacks are not fully strategic like other models explored in the literature \cite{chen2018strategyproof}, we believe they approximate worst-case perturbations made by an agent without knowledge of the neural network gradient, but with access to changing a small number of features. 

\subsection{Robust Optimization Algorithm} \label{robust-opt}

\begin{algorithm}[t!]
   \caption{Oracle Efficient Improper Robust Optimization \cite{chen2017robust}}
   \label{alg:MWU}
\begin{algorithmic}[1]
   \STATE {\bfseries Input:} Objectives $\mathcal{L}=\{L_1,\dots,L_m\}$, Apx Bayesian Oracle $M$, parameters $T,\eta$
   \FOR{$t=1$ {\bfseries to} $T$}
   \STATE $w_t[i] \propto \exp \eta \left( \sum_{\tau=1}^{t-1} L_i(h_\tau) \right)$\\
   \STATE $h_t=M(w_t)$\\
   \ENDFOR
   \STATE {\bfseries Output:} uniform distribution over $\{h_1,\dots,h_T\}$

\end{algorithmic}
\end{algorithm}

To make our loan grade prediction classifier robust against these perturbations, we train our model with the robust optimization algorithm proposed by Chen et al. \cite{chen2017robust}. The key idea of the algorithm is that, given a model (e.g. a neural network) and a set of perturbations, the algorithm can identify which perturbations have high classification losses, and focus more on those perturbation types during the training. 

More formally, given a distribution $D=(D_1,\dots,D_m)$ over loss functions $\mathcal{L}=\{L_1,\dots,L_m\}$, we would like to minimize the loss in the worst-case over $\mathcal{L}$, namely $\tau=\min_{h \in \mathcal{H}}\max_{i \in [m]} L_i(h)$. We call this the bottleneck loss throughout the rest of this paper. 

Moreover, we assume that we are given an $\alpha$-approximate Bayesian oracle $M$. That is, $M(D)$ can find a hypothesis $h^*$ that $\alpha$-approximates the minimum expected risk
$$\E_{L \sim D}[L(h^*)] \leq \alpha \min_{h \in \mathcal{H}}\E_{L \sim D}[L(h)]$$
In our setting, $L$ corresponds to the loss of the model over different perturbations. Furthermore, we assume that the neural network trained over the perturbations is an $\alpha$-approximate Bayesian oracle.

The algorithm for optimizing this objective can be thought of as a no-regret dynamics on a zero-sum game. Concretely, the algorithm runs a multiplicative weight update (MWU) over $\mathcal{L}$ (Algorithm~\ref{alg:MWU}). Intuitively, the algorithm first assigns uniform importance weights to all perturbation types. Then in each iteration, it puts more weights to perturbations with higher loss.

Under this framework, Chen et al. \citep{chen2017robust} show that when $\eta = \sqrt{\log(m)/2T}$, the ensemble hypothesis $h^* = \frac{1}{T}\sum_{t=1}^{T} h_t$ satisfies
$$\max_{i \in [m]} L_i(h^*) \leq \alpha \min_{h \in \mathcal{H}} \max_{i \in [m]} L_i(h) + \sqrt{\frac{2 \log(m)}{T}}$$ 

In other words, the model can be robust against a distribution of perturbations, with the same $\alpha$ multiplicative factor loss, and an additive error that goes to zero as iterations increase.

%% file: attacks.tex
\section{Evaluation of Perturbations}

\begin{figure}[t!]
\centering
\includegraphics[width=\linewidth]{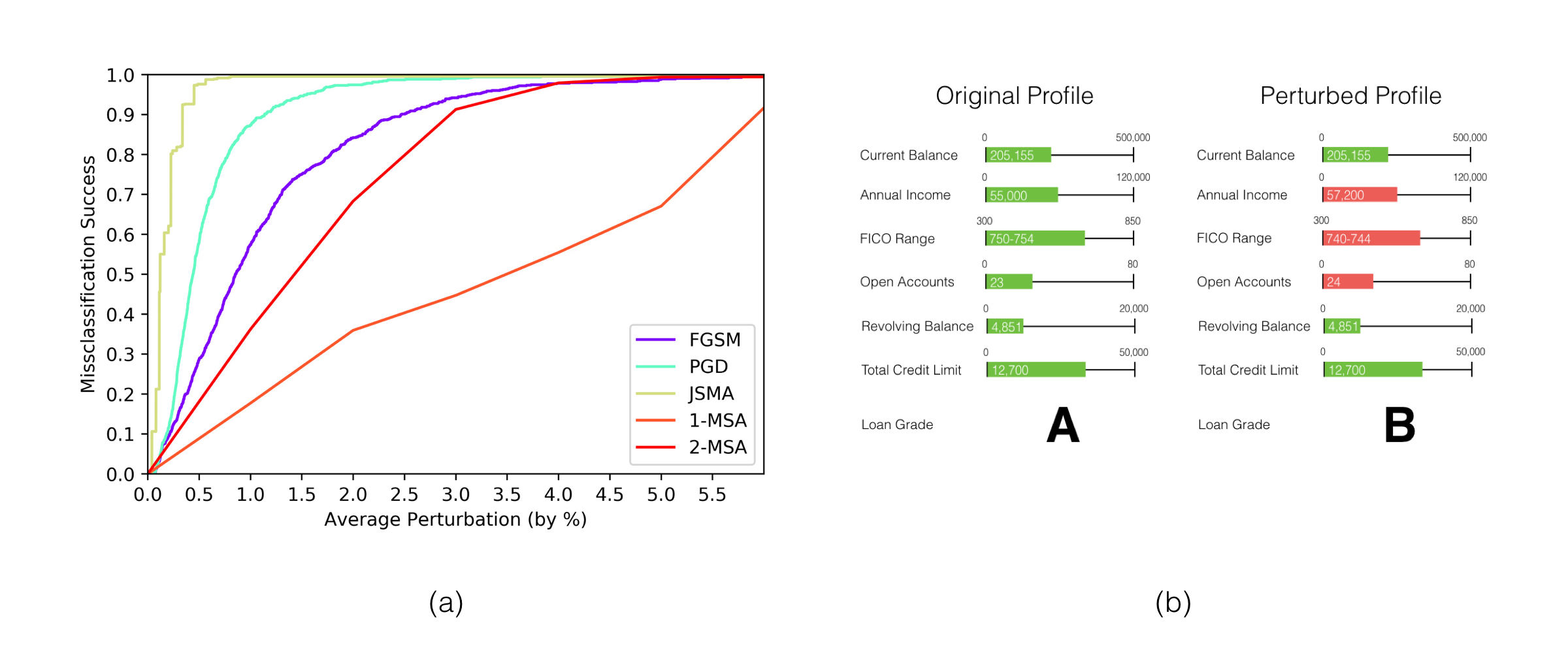} 
\caption{(a): Perturbations against a pre-trained classification model. The horizontal axis is the average percentage change in the perturbed features. The vertical axis is the rate of originally correctly classified samples that were misclassified upon small perturbations. (b): Example of individual perturbations of up to $4\%$ on three inputs, which lead to a change in loan grade from A to B.}
\label{fig:IndividualAttacks}
\end{figure}

We implemented perturbation methods on our loan category prediction model, varying the perturbation size and evaluating the success of changing the predicted label. We evaluated the extent to which perturbed train examples were misclassified as the perturbation budget increased. 

We found that small perturbations were effective at perturbing data by small amounts to change loan categories (Figure~\ref{fig:IndividualAttacks}). In particular, JSMA demonstrated an ability to target specific features that significantly affected the neural network's loan category prediction.

FGSM, which is based on perturbing all features by a constant amount, was more inefficient with budgeting perturbation size compared to other methods. PGD likely achieved higher perturbation efficiency by not perturbing features with low salience as much.

1-MSA, which only changed loan-to-income ratio for applicants, performed quite well, despite its constraints. MSA did not utilize gradients of the neural networks, which put it at a distinct disadvantage compared to other networks. However, MSA demonstrates that even a small change in a single, preselected variable can easily impact loan category classification. A loan applicant could easily round off this input (by rounding either income or loan) and change their loan category. Meanwhile, 2-MSA, which only extends upon 1-MSA by perturbing one additional variable, achieves significantly better results. Overall, MSA suggests that traditional, gradient-based methods are not necessary to produce small perturbations for successful misclassifications.

%% file: classifier.tex
\section{Robust Classification}

\begin{figure}[t!]
\centering
\centering
\includegraphics[width=\linewidth]{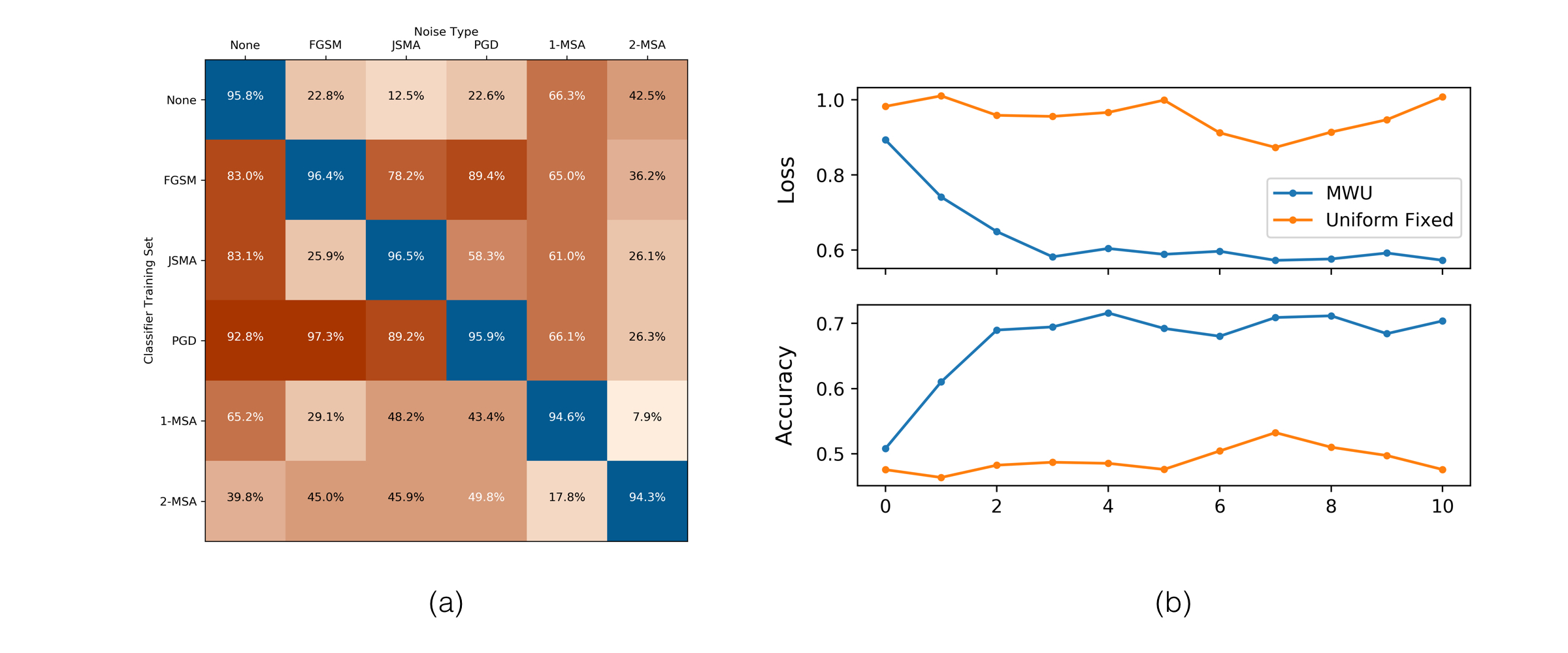}
\caption{(a): Payoff matrix representing the minmax problem being addressed by the robust optimization algorithm for loan classification. (b): Bottleneck loss (top) and accuracy (bottom) over 10 iterations of the robust optimization algorithm.}
\label{fig:mwu}
\end{figure}

We experimentally show that the robust optimization algorithm introduced in \cite{chen2017robust} improves the performance of our loan grade prediction model. Specifically, we train a neural network model with the same architecture as in Section~\ref{loan}. We train the model over the five perturbations introduced in Section~\ref{perturbations} as well as the original dataset with no perturbations (referred to as None).

As stated in Section~\ref{robust-opt}, the robust optimization framework can be thought of as a minmax game. In particular, it is a two player game where one player has classifiers as strategies and another player has perturbation types as strategies. The payoff matrix of the specific game played in this loan grade prediction optimization is shown in Figure~\ref{fig:mwu}a. It can easily be seen that if the row player commits to a classifier trained on a specific perturbed dataset, the column player can easily choose a different perturbation type to induce misclassification. This motivates the use of Algorithm~\ref{alg:MWU}, which is essentially estimating the optimal mixed strategy over the different classifiers for the row player.

The evaluation of the algorithm is based on the bottleneck loss as introduced in section~\ref{robust-opt}. The baseline is a training with fixed uniform distribution, meaning that we do not update the distribution over perturbation types at each iteration of the algorithm. This can be thought of as uniformly randomizing over the row strategies in Figure~\ref{fig:mwu}a.

The result of the robust optimization is shown in Figure~\ref{fig:mwu}b. The horizontal axis is the number of iterations ($T$ in Algorithm~\ref{alg:MWU}). For the top plot, the vertical axis is the bottleneck loss, and for the bottom plot, it is the accuracy corresponding to that loss. The robust optimization algorithm reduced the bottleneck loss from 1.0 to 0.6 and increased the bottleneck accuracy from 50\% to 70\%. The loan grade prediction model trained with the robust optimization algorithm is more robust against perturbations.

\section{Conclusion}
We introduce the first treatment, to our knowledge, of adversarial attacks and defenses on neural networks used in financial services. While small perturbations in input data can lead loan grade models to misclassify, building robustness into these classifiers can help ensure that models perform as expected on inputs that are slightly different from those the model was trained on. Robust training can ensure that a model will behave as specified, which helps protect against differing distributions of inputs or information manipulators.

Future work in this space would benefit from more exploration into the feasible set of changeable features and degrees of perturbation.  
By better understanding the set of feasible perturbations and cost of misreporting for individuals, robust classification could help make algorithmic systems build stronger decision boundaries across problem types. Furthermore, an analysis into whether weaker decision boundaries exist for different classes of individuals could help highlight issues of fairness or differential robustness. Finally, further exploration of non-laboratory domains across financial service areas would help better show how robustness can be employed in practice.